\pdfoutput=1

\documentclass[11pt]{article}

\usepackage{acl}

\usepackage{times}
\usepackage{latexsym}

\usepackage[T1]{fontenc}

\usepackage[utf8]{inputenc}

\usepackage{microtype}

\usepackage{inconsolata}
\usepackage{xspace}

\usepackage{graphicx}
\usepackage{amsmath}
\usepackage{pgfplots}
\usepackage{pgfplotstable}
\usepackage{booktabs}
\usepackage{multirow}

\usepackage{inconsolata}
\usepackage{graphicx}
\usepackage{hyperref}
\usepackage{booktabs}
\usepackage[svgnames]{xcolor}
\usepackage{colortbl}
\usepackage{tabularx}
\usepackage{multirow}
\usepackage{tikz}
\usepackage{pgfplots}
\pgfplotsset{compat=1.18} 
\definecolor{lightgray}{gray}{0.95}
\usepackage[framemethod=default]{mdframed}

\newcommand{\eg}{\emph{e.g.,}\xspace}

\title{
GeoPQA: Bridging the Visual Perception Gap in MLLMs for Geometric Reasoning}

\author{
  Guizhen Chen$^{1,2,}$\thanks{Guizhen is under the Joint PhD Program between Alibaba and NTU.}
  \quad Weiwen Xu$^{2}$
  \quad Hao Zhang$^{2}$
  \quad Hou Pong Chan$^{2}$\\
   \quad \textbf{Deli Zhao}$^{2,3}$
   \quad \textbf{Anh Tuan Luu}$^{1}$
   \quad \textbf{Yu Rong}$^{2,3}$ \\
  $^1$ Nanyang Technological University $^2$ DAMO Academy, Alibaba Group
  $^3$ Hupan Lab
}

\begin{document}
\maketitle
\begin{abstract}

Recent advancements in reinforcement learning (RL) have enhanced the reasoning abilities of large language models (LLMs), yet the impact on multimodal LLMs (MLLMs) is limited. Particularly in vision-intensive tasks like geometric reasoning, MLLMs hallucinate frequently, leading to inaccurate reasoning. We attribute this to the \textit{perceptual bottleneck} in MLLMs, which caps the benefits of reasoning training. To quantify this, we design a Geo-Perception Question-Answering (GeoPQA) benchmark, targeting basic geometric concepts and spatial relationships. Experiments on GeoPQA reveal significant shortcomings of MLLMs in visual perception, which constrain RL reward signals for effective training. To address this bottleneck, we propose a two-stage RL training framework by first enhancing the visual perception of geometric structures, then fostering reasoning capabilities. Applied to Qwen2.5-VL-3B-Instruct, our two-stage training improves geometric reasoning by 9.7\% and geometric problem solving by 9.1\%, compared to the direct reasoning training approach. Our method also generalizes to other vision-intensive domains like figure understanding, highlighting the importance of perceptual grounding in effective MLLM reasoning.\footnote{\url{https://github.com/DAMO-NLP-SG/GeoPQA}}

\end{abstract}

\section{Introduction}

\begin{figure}[t]
    \centering
    \includegraphics[trim={0.0cm 2.8cm 0.0cm 3.2cm},clip,width=1\linewidth]{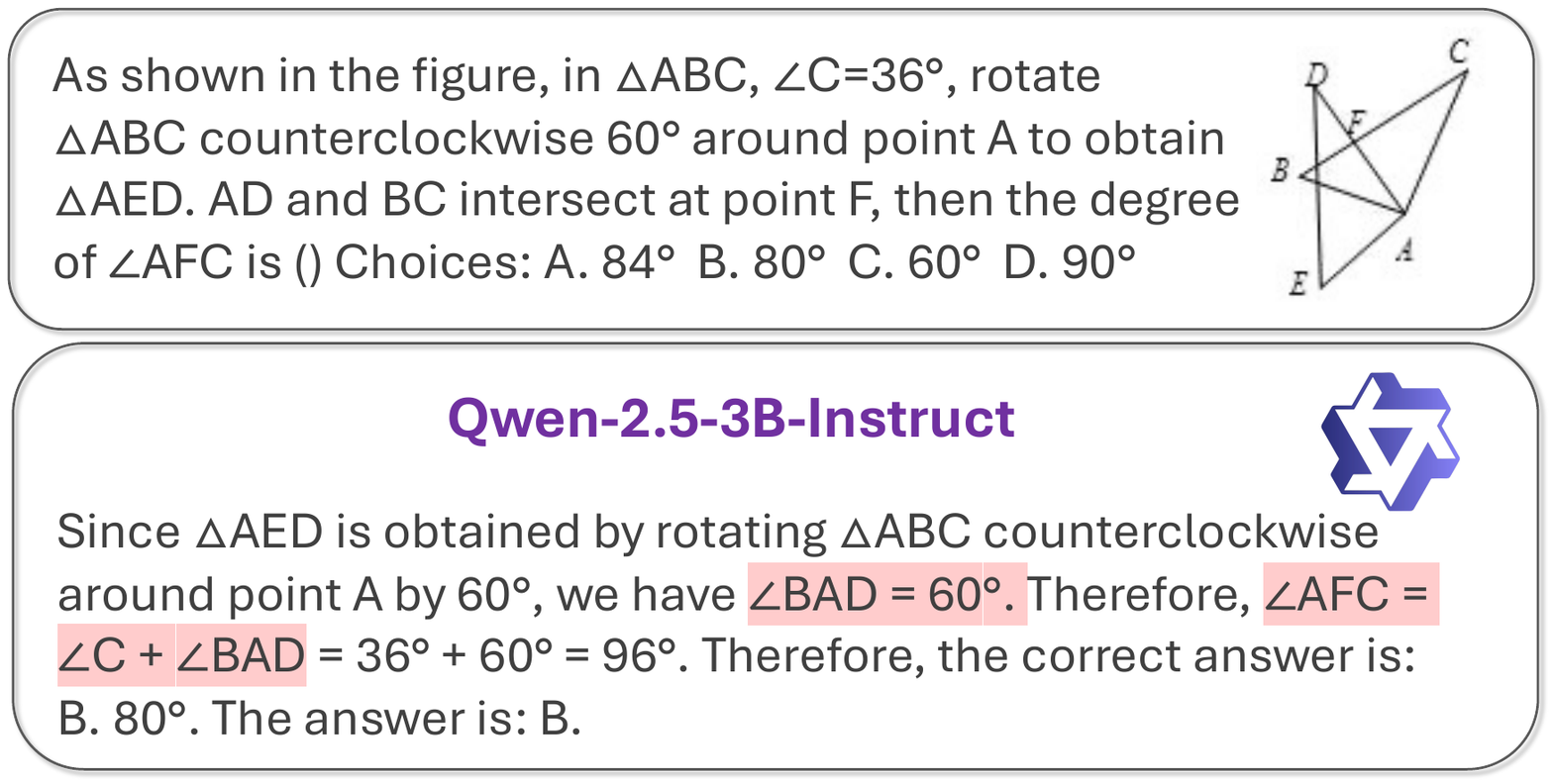}
    \caption{Illustration of how perceptual errors cascade into flawed reasoning in Qwen-2.5-3B-Insturct. The model misidentifies rotation angles and misinterprets angle composition.}
    \label{fig:perception-error}
\end{figure}

Recent advances in reasoning models such as DeepSeek-R1 \citep{deepseekai2025deepseekr1} have demonstrated that reinforcement learning with verifiable reward (RLVR) can markedly strengthen the reasoning abilities of large language models (LLMs; \citet{kimiteam2025kimik15,openai2024o1,openai2025o3}). Motivated by these successes, several studies have applied similar RL training recipes to multimodal LLMs (MLLMs; \citealt{seed15VL,yang2025r1onevision,shen2025vlm,kimiteam2025kimivl,yuan2025vl,xu2025lingshu,leng2025mmr1}). However, performance gains in vision-intensive reasoning benchmarks, such as MathVerse~\citep{zhang2024mathverse} and MathVista~\citep{lu2024mathvista}, remain relatively limited.
A closer examination suggests that these limitations often originate from more foundational issues in visual understanding, even before complex reasoning is attempted. Figure~\ref{fig:perception-error} shows an example of a model struggling with identifying the rotation angle, a task that is easy for humans. Such fundamental errors in vision understanding affect subsequent logical deductions, preventing the model from being rewarded during RL training.



We hypothesize that \textit{RL's efficacy in MLLMs is upper-bounded by their underlying visual perception ability}. Inadequate perception restricts the attainable reward signals and, consequently, the scope of reasoning improvement in RL training.
To quantify this perceptual bottleneck, we curate visual perception QAs to assess models' understanding of basic geometric concepts and relationships. Empirical results indicate that MLLMs frequently fail to perceive geometric information, in contrast to the near-perfect human performance.

\begin{figure}[t]
    \centering
    \includegraphics[trim={0.0cm 3cm 0.0cm 3.2cm},clip,width=0.95\linewidth]{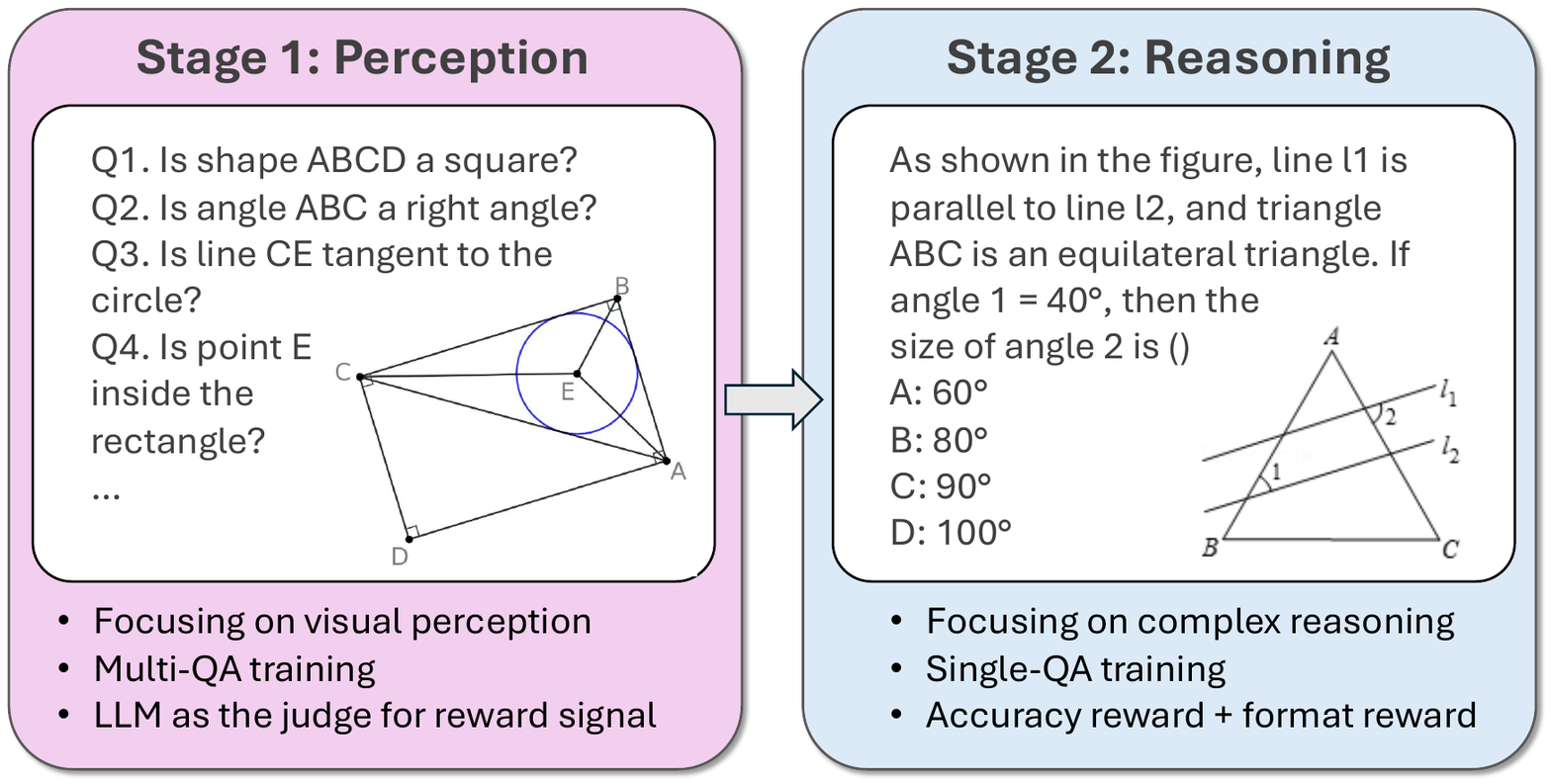}
    \caption{Overview of our two-stage RL framework. }
    \label{fig:framework}
\end{figure}

To overcome the perceptual bottleneck and promote effective reasoning training, we propose a two-stage RL framework comprising a perception stage followed by a reasoning stage, as shown in Figure \ref{fig:framework}. 
The first stage enhances models' visual understanding of basic geometric concepts and relationships using our curated perception-oriented QA dataset derived from both real and synthetic geometric diagrams. 
Building on the improved perceptual foundation, the second stage focuses on reasoning-oriented training, enabling the model to leverage its enhanced visual understanding and concentrate more effectively on the logical deduction process.
Experiments on MathVista show that our two-stage framework is superior to direct single-stage reasoning training, with a 9.7\% and 9.1\% accuracy improvement in geometric reasoning and problem solving, respectively. It also outperforms larger open-source MLLMs and previous mathematical visual specialist models. Beyond geometric tasks, we further demonstrate that the perception-first paradigm generalizes to other vision-intensive tasks like figure and textbook understanding.

Our main contributions are threefold: (1) We reveal and quantify the perceptual limitations of current MLLMs in geometric tasks through targeted perception QAs, which are often overlooked by approaches focused solely on reasoning. (2) We introduce a two-stage reinforcement learning framework that first enhances visual perception before training for complex reasoning. (3) We validate the effectiveness of our approach on challenging geometric reasoning benchmarks, outperforming direct reasoning training and demonstrating potential generalization to other vision-intensive tasks.

\section{Methodology}
In this section, we first systematically assess the perceptual capabilities of MLLMs in the geometric domain. Next, we develop a two-stage RL framework that first enhances perception and subsequently boosts reasoning capabilities of MLLMs.

\subsection{Preliminary Analysis
}
\label{sec:prelim}

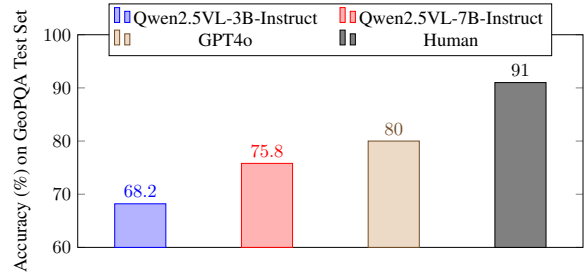
\begin{figure}[t!]
\centering
\resizebox{1.0\linewidth}{!}{
\begin{tikzpicture}
\begin{axis}[
    ybar,
    bar width=30pt,
    width=12cm,
    height=6cm,
    ymin=60,
    ymax=100,
    xmin=0.5, xmax=3,
    ylabel={Accuracy (\%) on GeoPQA Test Set},
    xtick=\empty, 
    enlarge x limits=0.2,
    legend style={
        at={(0.5,0.9)},
        anchor=south,
        legend columns=2,
        /tikz/every even column/.append style={column sep=1em}
    },
    nodes near coords,
    nodes near coords align={vertical},
]

\addplot coordinates {(1, 68.2)};
\addlegendentry{Qwen2.5VL-3B-Instruct}

\addplot coordinates {(1.5, 75.8)};
\addlegendentry{Qwen2.5VL-7B-Instruct}

\addplot coordinates {(2, 80.0)};
\addlegendentry{GPT4o}

\addplot coordinates {(2.5, 91.0)};
\addlegendentry{Human}

\end{axis}
\end{tikzpicture}
}
\caption{Perceptual capability of MLLMs against humans on GeoPQA test set. }
\label{fig:prelim}
\end{figure}

While existing research suggests that MLLMs struggle with geometric images, few studies comprehensively assess their perceptual abilities in this domain. MathVerse~\citep{zhang2024mathverse} infers perceptual ability based on models' capabilities to answer geometric questions with varying levels of visual descriptions in the text format. VisOnly~\citep{DBLP:journals/corr/VisOnlyQA} and Geoperception~\citep{zhang2024euclidsuperchargingmultimodalllms} directly evaluate geometric perception, but use a limited range of templates, hindering the evaluation of diverse perceptual skills.

To thoroughly evaluate models' geometric perception, we construct a test set using the image-caption pairs in the Geo170K dataset~\citep{DBLP:conf/iclr/Geo170k25}. Specifically, we prompt Gemini-2.0-Flash-Thinking~(Gemini-FT; \citet{Gemini2FlashThinking}) to generate questions that require recognizing basic visual elements and spatial relationships directly from the image descriptions (see Appendix \ref{appendix:prompt}). These questions cover: (1) \textbf{basic geometric elements} such as identifying shapes (\eg triangles, circles), comparing lengths, and recognizing angles (\eg right, acute, obtuse); (2) \textbf{geometric relationships} such as intersection, parallelism, perpendicularity, and tangency. To facilitate automatic evaluation, the answers are designed for easy verification, restricting them to yes/no, numerical values, or simple strings (\eg ``ABC''). 

We assess the performance of several representative MLLMs in Figure~\ref{fig:prelim}. The results
reveal significant deficiencies in models in answering these basic visual questions, even for state-of-the-art models like GPT4o, while humans easily attain over 90\% accuracy in these questions. This highlights a critical perceptual gap that limits the effectiveness of subsequent reasoning training via RL.

\subsection{Framework Overview}

To overcome the perceptual bottleneck, we introduce a two-stage RL framework: perception followed by reasoning. Stage 1 focuses on improving the model's ability to accurately perceive and interpret geometric information, while stage 2 leverages the enhanced perceptual foundation to develop more complex, multi-step reasoning capabilities.

\subsection{Stage 1: Perception-oriented Training}

\paragraph{Training data.}
To enhance the geometric perception of MLLMs, we curate a comprehensive \textbf{Geo-Perception Question-Answering (GeoPQA)} training dataset, comprised of both real-world and synthetic figures. For real-world images, we employed the same methodology as described in our preliminary study: leveraging Gemini-FT to generate perception-focused QAs. We further augment this data with synthetically generated geometric figures to cover a wider range of scenarios. Following methodologies in AlphaGeometry~\citep{alphaGeometry24} and AutoGen~\citep{DBLP:journals/corr/autogen}, we create basic shapes and composite shapes. Additionally, we use geometric annotations to visually enrich the diagrams (see Appendix~\ref{appendix:synthetic}).
The perception QAs for the synthetic figures are generated similarly to those for real images, concentrating on elements and relationships present in the generated diagrams.
Since each image contains rich visual information, we generate a set of perception questions $Q = (q_1, \ldots, q_n)$ per image ($n \le 7$). Dataset statistics are presented in Appendix \ref{appendix:data-stats}.

\paragraph{Quality control.}
To ensure quality, we use GPT-4o~\citep{openai2024gpt4o} to filter out questions where: (1) the image does not explicitly contain the information required to answer the question, or (2) the provided ground-truth answer contradicts the information evident in the image description. To validate the quality of the dataset after GPT-4o filtering, we perform a human inspection on 100 random samples: 92\% are valid and high-quality. More details are shown in Appendix \ref{appendix:quality}.

\paragraph{RL training.}
The input to the model at this stage is formulated as
\[x = (I, q_1, \ldots, q_n)\]
where $I$ is the instruction, and $q_i$ is a visual perception question. Given an input $x$, the policy $\pi_\theta$ generates a free-form textual response $y \sim \pi_\theta(y|x)$. This response $y$ is expected to contain the answers to all $n$ questions. Let the ground-truth answers be 
\[A = (a_1, \ldots, a_n)\]
where $a_i$ is the ground-truth answer to $q_i$. To evaluate the correctness of the generated response $y$, we employ GPT-4o-mini~\citep{openai2024gpt4o} as the judge, denoted as $J$. The judge $J$ parses the response $y$ to extract the model's predicted answers for each perception question, yielding $J(y) = \hat{A} = (\hat{a}_1, \ldots, \hat{a}_n)$. The accuracy reward $R(x, y)$ for a given input $x$ and generated response $y$ is defined as:
\begin{align*}
R(x, y) = 
\begin{cases}
1, & \text{if } \hat{a}_i = a_i, \forall i \in \{1, \ldots, n\}\\
0, & \text{otherwise}.
\end{cases}
\end{align*}
This strict reward function grants a positive reward only if all perception questions are answered correctly. To mitigate reward hacking (\eg the model learning to always output ``yes''), we downsample training instances where all ground-truth answers are ``yes''. Other training details are kept the same as in the original Group Relative Policy Optimization (GRPO; \citet{deepseekai2025deepseekr1}).

\subsection{Stage 2: Reasoning-oriented Training}

With the improved perceptual capabilities, the MLLM proceeds to stage 2, where it is trained on geometric reasoning tasks.
We use the QA tuning subset from Geo170K~\cite{DBLP:conf/iclr/Geo170k25}. We follow the standard GRPO setup to apply RL training at this stage.

\subsection{Implementation}
We train our models based on Qwen-2.5-VL-3B-Instruct and Qwen-2.5-VL-7B-Instruct~\cite{DBLP:journals/corr/qwen2.5vl}. Besides the backbone models, we compare against several baselines, including five proprietary MLLMs~\citep{bai2024qwenvl, geminiteam2025geminifamilyhighlycapable, 2023GPT4VisionSC, openai2024gpt4o} and six open-source MLLMs~\citep{SPHINX, DBLP:conf/iclr/Geo170k25, shi-etal-2024-math, zhang2025mavis, dong2024internlmxcomposer2masteringfreeformtextimage, chen2025expandingperformanceboundariesopensource}. We evaluate models on geometry reasoning (GR) and geometry problem-solving (GPS) on MathVista~\citep{lu2024mathvista}. Training and evaluation details are provided in Appendix~\ref{appendix:training} and \ref{appendix:evaluation} respectively. 

\begin{table}[ht]
\centering
\resizebox{1.0\linewidth}{!}{
\begin{tabular}{lcc}
\toprule
\textbf{Model} & \textbf{GR} & \textbf{GPS} \\
\midrule
\multicolumn{3}{c}{Proprietary MLLMs} \\
\midrule
Qwen-VL-Plus & 39.3 & 38.5 \\
GeminiPro & -- & 40.4 \\
GPT-4V & 51.0 & 50.5 \\
GeminiUltra & -- & 56.3 \\
GPT-4o & \textbf{74.1} & \textbf{75.0} \\
\midrule
\multicolumn{3}{c}{Open-source MLLMs}\\
\midrule
SPHINX-MoE (8 $\times$ 7B) & 30.5 & 31.2\\
G-LLAVA* (13B) & -- & 56.7 \\
Math-LLAVA* (13B) & 56.5 & 57.7 \\
MAVIS* (7B) & 63.2 & 64.1 \\
InternLM-XC2 (7B) &  62.3 & 63.0 \\
InternVL2.5 (4B) & 64.4 & 67.3 \\
\midrule
Qwen2.5-VL-3B-Instruct & 63.2 & 63.9 \\
\enspace w/ Reasoning & 62.3 & 63.0 \\
\enspace w/ Perception and Reasoning mixed  & 65.7 & 65.9\\
\enspace w/ Perception followed by Reasoning & \textbf{72.0} & \textbf{72.1} \\
\bottomrule
\end{tabular}
}
\caption{Results of MathVista-\textit{testmini}~\citep{lu2024mathvista} on geometry reasoning (GR) and geometry problem solving (GPS). * denotes visual specialists in mathematics. The highest scores for proprietary and open-source MLLMs are \textbf{bolded}.}
\label{tab:main}
\end{table}

\section{Results and Analyses}

\paragraph{Main results.} 
Table \ref{tab:main} shows that our two-stage approach significantly outperforms the reasoning-only training approach by a large margin of 9.7\% in GR and 9.1\% in GPS. Notably, the reasoning-only approach scores even slightly lower than the original baseline. This suggests that directly applying RL for reasoning without addressing underlying perceptual limitations can be ineffective or even detrimental, and our perception training effectively bridges the perception gap.

In addition, we observe that incorporating perception data, whether mixed or sequential, boosts reasoning performance over reasoning-only RL. This validates the effectiveness of our perception QA dataset. Moreover, a structured, sequential approach of perception followed by reasoning training yields greater benefits than simply mixing perception and reasoning data, validating the effectiveness of our two-stage framework.

Compared to other leading open-source MLLMs and models specialized in mathematics, our method establishes new state-of-the-art performance on geometric tasks, even with a much smaller model size. While GPT-4o remains the top performer among proprietary models with 74.1\% on GR and 75.0\% on GPS, our two-stage framework, applied to the Qwen2.5-VL-3B-Instruct, considerably narrows the performance gap to just around 2\%.

\paragraph{Enhancement on visual perception.}
Table \ref{tab:geopqa} directly quantifies the benefits of our approach on models' visual perception, measured by GeoPQA. The results provide several key insights:
\begin{itemize}
    \item \textbf{Effectiveness of perception training:} Perception training significantly improves the performance on GeoPQA by 21.6\%, which validates its effectiveness in directly enhancing geometric visual perception.
    \item \textbf{Necessity of a staged approach:} Reasoning training degrades performance on GeoPQA by 15.1\%, suggesting that training directly on high-level reasoning can cause the model to neglect or unlearn basic perceptual abilities, which justifies our two-stage approach.
    \item \textbf{Balanced two-stage approach:} Our two-stage approach maintains a high perception score of 83.2\%, with a 15\% gain over the baseline while also achieving the significant reasoning gains reported in our main results.
\end{itemize}

\begin{table}[t]
\centering
\resizebox{1.0\linewidth}{!}{
    \begin{tabular}{lc}
    \toprule
    \textbf{Model} & \textbf{GeoPQA} \\
    \midrule
    \rowcolor{lightgray} Qwen2.5-VL-3B-Instruct & 68.2 \\
    \enspace w/ Perception & 89.8\\
    \enspace w/ Reasoning & 53.1 \\
    \enspace w/ Perception followed by Reasoning & 83.2 \\
    \bottomrule
    \end{tabular}
}
\caption{Performance of Qwen2.5-VL-3B-Instruct on GeoPQA.}
\label{tab:geopqa}
\end{table}

\begin{table}[ht]
\centering
\resizebox{1.0\linewidth}{!}{
\begin{tabular}{lccccc}
\toprule
\textbf{Model} & \textbf{TD} & \textbf{TL} & \textbf{VI} & \textbf{VD} & \textbf{VO}\\
\midrule
\rowcolor{lightgray} Qwen2.5-VL-3B-Instruct & 46.5
&39.6
&37.5
&38.4
&37.1
 \\
\enspace w/ Reasoning & 49.8
&44.3
&38.0
&41.8
&43.9
 \\
\enspace w/ Perception and Reasoning mixed  & 56.1
&51.6
&48.6
&48.2
&39.8
\\
\enspace w/ Perception followed by Reasoning & 55.3
&52.5
&47.5
&47.6
&45.5
 \\
\bottomrule
\end{tabular}
}
\caption{Results of MathVerse-\textit{testmini}~\citep{lu2024mathvista} on the plane geometry subset.}
\label{tab:mathverse}
\end{table}

\paragraph{Impact of training strategy on tasks of different vision intensity.} 
To further understand the benefits of our training framework, we analyze its performance on MathVerse~\citep{zhang2024mathverse} Plane Geometry problems, which includes 5 vision intensities: Text Dominant (TD), Text Lite (TL), Vision Intensive (VI), Vision Dominant (VD), and Vision Only (VO). As shown in Table~\ref{tab:mathverse}, across all categories, both perception-involved methods generally outperform the reasoning-only approach and the base model. Notably, our two-stage approach excels in the Vision Only scenario compared to the single-stage mixed approach. 
The results suggest that while mixing perception and reasoning data can be beneficial, a dedicated initial stage focused purely on perception, as in our two-stage framework, is crucial for tasks where the model cannot rely on textual cues to compensate for perceptual weaknesses.

\paragraph{Impact of multiple perception QAs per image.} 
In stage 1, we formulate training samples by concatenating multiple perception questions for a single image. To evaluate the effectiveness of this setting, we conduct an ablation study comparing it against the conventional method, where each perception question is treated as an individual training sample. The results in Table~\ref{tab:analysis-multiple} show that training with multiple QAs per image demonstrates a substantial advantage for downstream reasoning tasks, with an improvement of 9.6\% for GR and 10.6\% for GPS. While the multi-QA setup exhibits slightly lower performance on our perception task, this is potentially attributable to the stricter reward in the multi-QA training, in which the model only receives a positive reward if all sub-questions associated with an image are answered correctly. This more demanding training setup, however, encourages the model to learn more robustly on perception, ultimately leading to superior performance on downstream reasoning tasks.

\begin{table}[t]
\small
\centering
\resizebox{0.7\linewidth}{!}{
\begin{tabular}{lccc}
\toprule
\textbf{Training} & \textbf{GR} & \textbf{GPS} & \textbf{GeoPQA} \\
\midrule
Single QA & 52.7 & 52.9 & 95.4 \\
Multiple QAs & 62.3 & 63.5 & 94.0 \\
\bottomrule
\end{tabular}
}
\caption{Effects of perception training with single QA per image vs. multiple QAs per image.}
\label{tab:analysis-multiple}
\vspace{-0.15in}
\end{table}

\paragraph{Results on a larger scale.}
To demonstrate that our method remains effective at a larger scale, we extend our experiments to a 7B model. The results in Table \ref{tab:7b} are consistent with our observations with the 3B model, with 2.6\% gain in GR and 4.8\% improvement in GPS. Notably, our 7B model surpasses all other models, including the strong proprietary baseline GPT-4o (74.1\% GR, 75.0\% GPS).
These results reinforce our central claim: even for a more capable base model, enhancing foundational visual perception is a critical prerequisite for unlocking further gains in high-level reasoning and problem-solving.

\begin{table}[ht]
\centering
\resizebox{1.0\linewidth}{!}{
\begin{tabular}{lcc}
\toprule
\textbf{Model} & \textbf{GR} & \textbf{GPS}\\
\midrule
\rowcolor{lightgray} Qwen2.5-VL-7B-Instruct & 74.1 & 75.5 \\
\enspace w/ Reasoning & 73.6 & 75.0 \\
\enspace w/ Perception followed by Reasoning & \textbf{76.2} & \textbf{79.8} \\
\bottomrule
\end{tabular}
}
\caption{Performance of Qwen2.5-VL-7B-Instruct on geometry reasoning (GR) and geometry problem solving (GPS) from MathVista-\textit{testmini}~\citep{lu2024mathvista}.}
\label{tab:7b}
\end{table}

\paragraph{Generalization to other tasks.} 
To assess the broader impact of our two-stage training, we evaluate a diverse set of other tasks from MathVista, comparing our perception-then-reasoning approach against the reasoning-only baseline. The results are presented in Appendix~\ref{sec:othertasks}. 
Performance gains are observed in visually grounded tasks, including figure question answering (+1.5\%), textbook QA (+2.6\%), and scientific reasoning (+2.5\%), indicating that improved visual perception from stage 1 training facilitates more effective reasoning for tasks that involve interpreting diagrams.
The impact of geometry-focused perception training is less pronounced or slightly negative on tasks that are more text-reliant or require different types of visual understanding than geometry, such as numerical commonsense (-2.8\%) and math word problems (-1.1\%).

\section{Conclusion}
We investigate the fundamental challenge of improving geometric reasoning capabilities in MLLMs. Our analysis reveals that the effectiveness of reinforcement learning for reasoning is significantly constrained by MLLMs' visual perception, a critical bottleneck that is often not directly measured in previous work. By developing a targeted assessment of geometric perception and introducing a two-stage RL framework that explicitly enhances visual perception prior to reasoning training, we achieved substantial improvements on challenging benchmarks. The success of our perception-first training approach highlights an important principle for future work in multimodal reasoning: strong perceptual foundations are prerequisites for effective higher-level reasoning. Future directions include exploring whether our approach can enhance the performance of recent thinking-with-images approaches~\cite{openai2025o3,su2025thinkingimagesmultimodalreasoning,ma2024visaidmath} and generalizing our framework to other vision-intensive tasks like chart understanding~\cite{huang2024pixels}.

\section*{Limitations}
In our experiments, the accuracy of stage 1 training relies on an LLM judge (GPT-4o-mini) to parse free-form answers and determine the correctness for the perception QAs. This introduces extra cost and time calling the APIs.

\section*{Acknowledgment}
This research is jointly supported by DSO grant DSOCL23216, DAMO Academy Research Intern Program, and Alibaba-NTU Singapore Joint Research Institute. We would also like to extend our gratitude to Interdisciplinary Graduate Programme
and College of Computing and Data Science of NTU, for their support. We acknowledge the use of AI assistants (Google's Gemini, ChatGPT) for language editing and to improve the clarity of this manuscript.

\bibliography{custom}

\begin{thebibliography}{32}
\providecommand{\natexlab}[1]{#1}

\bibitem[{Bai et~al.(2024)Bai, Bai, Yang, Wang, Tan, Wang, Lin, Zhou, and Zhou}]{bai2024qwenvl}
Jinze Bai, Shuai Bai, Shusheng Yang, Shijie Wang, Sinan Tan, Peng Wang, Junyang Lin, Chang Zhou, and Jingren Zhou. 2024.
\newblock \href {https://openreview.net/forum?id=qrGjFJVl3m} {Qwen-{VL}: A versatile vision-language model for understanding, localization, text reading, and beyond}.

\bibitem[{Bai et~al.(2025)Bai, Chen, Liu, Wang, Ge, Song, Dang, Wang, Wang, Tang, Zhong, Zhu, Yang, Li, Wan, Wang, Ding, Fu, Xu, Ye, Zhang, Xie, Cheng, Zhang, Yang, Xu, and Lin}]{DBLP:journals/corr/qwen2.5vl}
Shuai Bai, Keqin Chen, Xuejing Liu, Jialin Wang, Wenbin Ge, Sibo Song, Kai Dang, Peng Wang, Shijie Wang, Jun Tang, Humen Zhong, Yuanzhi Zhu, Ming{-}Hsuan Yang, Zhaohai Li, Jianqiang Wan, Pengfei Wang, Wei Ding, Zheren Fu, Yiheng Xu, and 8 others. 2025.
\newblock \href {https://doi.org/10.48550/ARXIV.2502.13923} {Qwen2.5-vl technical report}.
\newblock \emph{CoRR}, abs/2502.13923.

\bibitem[{Chen et~al.(2025)Chen, Wang, Cao, Liu, Gao, Cui, Zhu, Ye, Tian, Liu, Gu, Wang, Li, Ren, Chen, Luo, Wang, Jiang, Wang, He, Shi, Zhang, Lv, Wang, Shao, Chu, Tu, He, Wu, Deng, Ge, Chen, Zhang, Wang, Dou, Lu, Zhu, Lu, Lin, Qiao, Dai, and Wang}]{chen2025expandingperformanceboundariesopensource}
Zhe Chen, Weiyun Wang, Yue Cao, Yangzhou Liu, Zhangwei Gao, Erfei Cui, Jinguo Zhu, Shenglong Ye, Hao Tian, Zhaoyang Liu, Lixin Gu, Xuehui Wang, Qingyun Li, Yimin Ren, Zixuan Chen, Jiapeng Luo, Jiahao Wang, Tan Jiang, Bo~Wang, and 23 others. 2025.
\newblock \href {https://arxiv.org/abs/2412.05271} {Expanding performance boundaries of open-source multimodal models with model, data, and test-time scaling}.
\newblock \emph{Preprint}, arXiv:2412.05271.

\bibitem[{DeepMind(2025)}]{Gemini2FlashThinking}
Google DeepMind. 2025.
\newblock \href {https://deepmind.google/technologies/gemini/flash-thinking/} {{Introducing Gemini 2.0: our new AI model for the agentic era}}.
\newblock Blog post on Google DeepMind website.

\bibitem[{DeepSeek-AI(2025)}]{deepseekai2025deepseekr1}
DeepSeek-AI. 2025.
\newblock \href {https://arxiv.org/abs/2501.12948} {Deepseek-r1: Incentivizing reasoning capability in llms via reinforcement learning}.
\newblock \emph{CoRR}, abs/2501.12948.

\bibitem[{Dong et~al.(2024)Dong, Zhang, Zang, Cao, Wang, Ouyang, Wei, Zhang, Duan, Cao, Zhang, Li, Yan, Gao, Zhang, Li, Li, Chen, He, Zhang, Qiao, Lin, and Wang}]{dong2024internlmxcomposer2masteringfreeformtextimage}
Xiaoyi Dong, Pan Zhang, Yuhang Zang, Yuhang Cao, Bin Wang, Linke Ouyang, Xilin Wei, Songyang Zhang, Haodong Duan, Maosong Cao, Wenwei Zhang, Yining Li, Hang Yan, Yang Gao, Xinyue Zhang, Wei Li, Jingwen Li, Kai Chen, Conghui He, and 4 others. 2024.
\newblock \href {https://arxiv.org/abs/2401.16420} {Internlm-xcomposer2: Mastering free-form text-image composition and comprehension in vision-language large model}.
\newblock \emph{Preprint}, arXiv:2401.16420.

\bibitem[{Gao et~al.(2025)Gao, Pi, Zhang, Ye, Zhong, Wang, Hong, Han, Xu, Li, and Kong}]{DBLP:conf/iclr/Geo170k25}
Jiahui Gao, Renjie Pi, Jipeng Zhang, Jiacheng Ye, Wanjun Zhong, Yufei Wang, Lanqing Hong, Jianhua Han, Hang Xu, Zhenguo Li, and Lingpeng Kong. 2025.
\newblock \href {https://openreview.net/forum?id=px1674Wp3C} {G-llava: Solving geometric problem with multi-modal large language model}.
\newblock In \emph{The Thirteenth International Conference on Learning Representations, {ICLR} 2025, Singapore, April 24-28, 2025}. OpenReview.net.

\bibitem[{Guo et~al.(2025)Guo, Wu, Zhu, Leng, Shi, Chen, Fan, Wang, Jiang, Wang et~al.}]{seed15VL}
Dong Guo, Faming Wu, Feida Zhu, Fuxing Leng, Guang Shi, Haobin Chen, Haoqi Fan, Jian Wang, Jianyu Jiang, Jiawei Wang, and 1 others. 2025.
\newblock Seed1. 5-vl technical report.
\newblock \emph{arXiv preprint arXiv:2505.07062}.

\bibitem[{Huang et~al.(2024)Huang, Chan, Fung, Qiu, Zhou, Joty, Chang, and Ji}]{huang2024pixels}
Kung-Hsiang Huang, Hou~Pong Chan, Yi~R Fung, Haoyi Qiu, Mingyang Zhou, Shafiq Joty, Shih-Fu Chang, and Heng Ji. 2024.
\newblock From pixels to insights: A survey on automatic chart understanding in the era of large foundation models.
\newblock \emph{IEEE Transactions on Knowledge and Data Engineering}.

\bibitem[{Kamoi et~al.(2024)Kamoi, Zhang, Das, Zhang, and Zhang}]{DBLP:journals/corr/VisOnlyQA}
Ryo Kamoi, Yusen Zhang, Sarkar Snigdha~Sarathi Das, Ranran~Haoran Zhang, and Rui Zhang. 2024.
\newblock \href {https://doi.org/10.48550/ARXIV.2412.00947} {Visonlyqa: Large vision language models still struggle with visual perception of geometric information}.
\newblock \emph{CoRR}, abs/2412.00947.

\bibitem[{Leng et~al.(2025)Leng, Wang, Li, Zhang, Hu, Zhang, Zhang, Jiang, Li, Zhao, Wang, Rong, Sun, and Lu}]{leng2025mmr1}
Sicong Leng, Jing Wang, Jiaxi Li, Hao Zhang, Zhiqiang Hu, Boqiang Zhang, Hang Zhang, Yuming Jiang, Xin Li, Deli Zhao, Fan Wang, Yu~Rong, Aixin Sun, and Shijian Lu. 2025.
\newblock Mmr1: Advancing the frontiers of multimodal reasoning.
\newblock \url{https://github.com/LengSicong/MMR1}.

\bibitem[{Liu et~al.(2024)Liu, Zhang, Qiu, Huang, Lin, Zhao, Geng, Lin, Jin, Zhang, Shao, Xu, He, He, Shao, Lu, Qiao, Li, and Gao}]{SPHINX}
Dongyang Liu, Renrui Zhang, Longtian Qiu, Siyuan Huang, Weifeng Lin, Shitian Zhao, Shijie Geng, Ziyi Lin, Peng Jin, Kaipeng Zhang, Wenqi Shao, Chao Xu, Conghui He, Junjun He, Hao Shao, Pan Lu, Yu~Qiao, Hongsheng Li, and Peng Gao. 2024.
\newblock \href {https://proceedings.mlr.press/v235/liu24cc.html} {{SPHINX}-x: Scaling data and parameters for a family of multi-modal large language models}.
\newblock In \emph{Proceedings of the 41st International Conference on Machine Learning}, volume 235 of \emph{Proceedings of Machine Learning Research}, pages 32400--32420. PMLR.

\bibitem[{Lu et~al.(2024)Lu, Bansal, Xia, Liu, Li, Hajishirzi, Cheng, Chang, Galley, and Gao}]{lu2024mathvista}
Pan Lu, Hritik Bansal, Tony Xia, Jiacheng Liu, Chunyuan Li, Hannaneh Hajishirzi, Hao Cheng, Kai-Wei Chang, Michel Galley, and Jianfeng Gao. 2024.
\newblock \href {https://openreview.net/forum?id=KUNzEQMWU7} {Mathvista: Evaluating mathematical reasoning of foundation models in visual contexts}.
\newblock In \emph{The Twelfth International Conference on Learning Representations}.

\bibitem[{Ma et~al.(2024)Ma, Zhan, Wong, Li, Sun, Chan, and Chao}]{ma2024visaidmath}
Jingkun Ma, Runzhe Zhan, Derek~F. Wong, Yang Li, Di~Sun, Hou~Pong Chan, and Lidia~S. Chao. 2024.
\newblock \href {https://arxiv.org/abs/2410.22995} {Visaidmath: Benchmarking visual-aided mathematical reasoning}.
\newblock \emph{Preprint}, arXiv:2410.22995.

\bibitem[{OpenAI(2023)}]{2023GPT4VisionSC}
OpenAI. 2023.
\newblock \href {https://openai.com/index/gpt-4v-system-card/} {Gpt-4v(ision) system card}.

\bibitem[{OpenAI(2024{\natexlab{a}})}]{openai2024gpt4o}
OpenAI. 2024{\natexlab{a}}.
\newblock {GPT-4o}.
\newblock \url{https://openai.com/gpt-4o}.
\newblock Accessed: 2024-05-20.

\bibitem[{OpenAI(2024{\natexlab{b}})}]{openai2024o1}
OpenAI. 2024{\natexlab{b}}.
\newblock \href {https://arxiv.org/abs/2412.16720} {Openai o1 system card}.
\newblock \emph{CoRR}, abs/2412.16720.

\bibitem[{OpenAI(2025)}]{openai2025o3}
OpenAI. 2025.
\newblock \href {https://openai.com/index/introducing-o3-and-o4-mini/} {Introducing \textit{o3} and \textit{o4-mini}}.

\bibitem[{Shen et~al.(2025)Shen, Liu, Li, Fang, Ma, Liao, Shen, Zhang, Zhao, Zhang, Xu, and Zhao}]{shen2025vlm}
Haozhan Shen, Peng Liu, Jingcheng Li, Chunxin Fang, Yibo Ma, Jiajia Liao, Qiaoli Shen, Zilun Zhang, Kangjia Zhao, Qianqian Zhang, Ruochen Xu, and Tiancheng Zhao. 2025.
\newblock Vlm-r1: A stable and generalizable r1-style large vision-language model.
\newblock \emph{arXiv preprint arXiv:2504.07615}.

\bibitem[{Shi et~al.(2024)Shi, Hu, Bin, Liu, Yang, Ng, Bing, and Lee}]{shi-etal-2024-math}
Wenhao Shi, Zhiqiang Hu, Yi~Bin, Junhua Liu, Yang Yang, See-Kiong Ng, Lidong Bing, and Roy Ka-Wei Lee. 2024.
\newblock \href {https://doi.org/10.18653/v1/2024.findings-emnlp.268} {Math-{LL}a{VA}: Bootstrapping mathematical reasoning for multimodal large language models}.
\newblock In \emph{Findings of the Association for Computational Linguistics: EMNLP 2024}, pages 4663--4680, Miami, Florida, USA. Association for Computational Linguistics.

\bibitem[{Su et~al.(2025)Su, Xia, Guo, Liu, Ma, Qu, Liu, Li, Zeng, Yang, Li, Cheng, Ji, He, and Fung}]{su2025thinkingimagesmultimodalreasoning}
Zhaochen Su, Peng Xia, Hangyu Guo, Zhenhua Liu, Yan Ma, Xiaoye Qu, Jiaqi Liu, Yanshu Li, Kaide Zeng, Zhengyuan Yang, Linjie Li, Yu~Cheng, Heng Ji, Junxian He, and Yi~R. Fung. 2025.
\newblock \href {https://arxiv.org/abs/2506.23918} {Thinking with images for multimodal reasoning: Foundations, methods, and future frontiers}.
\newblock \emph{Preprint}, arXiv:2506.23918.

\bibitem[{Team et~al.(2025{\natexlab{a}})Team, Anil, Borgeaud, Alayrac, Yu, Soricut, Schalkwyk, Dai, Hauth, Millican, Silver, Johnson, Antonoglou, Schrittwieser, Glaese, Chen, Pitler, Lillicrap, Lazaridou, Firat, Molloy, Isard, Barham, Hennigan, Lee, Viola, Reynolds, Xu, Doherty, Collins, Meyer, Rutherford, Moreira, Ayoub, Goel, Krawczyk, Du, Chi, Cheng, Ni, Shah, Kane, Chan, Faruqui, Severyn, Lin, Li, Cheng, Ittycheriah, Mahdieh, Chen, Sun, Tran, Bagri, Lakshminarayanan, Liu, Orban, Güra, Zhou, Song, Boffy, Ganapathy, Zheng, Choe, Ágoston Weisz, Zhu, Lu, Gopal, Kahn, Kula, Pitman, Shah, Taropa, Merey, Baeuml, Chen, Shafey, Zhang, Sercinoglu, Tucker, Piqueras, Krikun, Barr, Savinov, Danihelka, Roelofs, White, Andreassen, von Glehn, Yagati, Kazemi, Gonzalez, Khalman, Sygnowski, Frechette, Smith, Culp, Proleev, Luan, Chen, Lottes, Schucher, Lebron, Rrustemi, Clay, Crone, Kocisky, Zhao, Perz, Yu, Howard, Bloniarz, Rae, Lu, Sifre, Maggioni, Alcober, Garrette, Barnes, Thakoor, Austin, Barth-Maron, Wong, Joshi,
  Chaabouni, Fatiha, Ahuja, Tomar, Senter, Chadwick, Kornakov, Attaluri, Iturrate, Liu, Li, Cogan, Chen, Jia, Gu, Zhang, Grimstad, Hartman, Garcia, Pillai, Devlin, Laskin, de~Las~Casas, Valter, Tao, Blanco, Badia, Reitter, Chen, Brennan, Rivera, Brin, Iqbal, Surita, Labanowski, Rao, Winkler, Parisotto, Gu, Olszewska, Addanki, Miech, Louis, Teplyashin, Brown, Catt, Balaguer, Xiang, Wang, Ashwood, Briukhov, Webson, Ganapathy, Sanghavi, Kannan, Chang, Stjerngren, Djolonga, Sun, Bapna, Aitchison, Pejman, Michalewski, Yu, Wang, Love, Ahn, Bloxwich, Han, Humphreys, Sellam, Bradbury, Godbole, Samangooei, Damoc, Kaskasoli, Arnold, Vasudevan, Agrawal, Riesa, Lepikhin, Tanburn, Srinivasan, Lim, Hodkinson, Shyam, Ferret, Hand, Garg, Paine, Li, Li, Giang, Neitz, Abbas, York, Reid, Cole, Chowdhery, Das, Rogozińska, Nikolaev, Sprechmann, Nado, Zilka, Prost, He, Monteiro, Mishra, Welty, Newlan, Jia, Allamanis, Hu, de~Liedekerke, Gilmer, Saroufim, Rijhwani, Hou, Shrivastava, Baddepudi, Goldin, Ozturel, Cassirer, Xu, Sohn,
  Sachan, Amplayo, Swanson, Petrova, Narayan, Guez, Brahma, Landon, Patel, Zhao, Villela, Wang, Jia, Rahtz, Giménez, Yeung, Keeling, Georgiev, Mincu, Wu, Haykal, Saputro, Vodrahalli, Qin, Cankara, Sharma, Fernando, Hawkins, Neyshabur, Kim, Hutter, Agrawal, Castro-Ros, van~den Driessche, Wang, Yang, yiin Chang, Komarek, McIlroy, Lučić, Zhang, Farhan, Sharman, Natsev, Michel, Bansal, Qiao, Cao, Shakeri, Butterfield, Chung, Rubenstein, Agrawal, Mensch, Soparkar, Lenc, Chung, Pope, Maggiore, Kay, Jhakra, Wang, Maynez, Phuong, Tobin, Tacchetti, Trebacz, Robinson, Katariya, Riedel, Bailey, Xiao, Ghelani, Aroyo, Slone, Houlsby, Xiong, Yang, Gribovskaya, Adler, Wirth, Lee, Li, Kagohara, Pavagadhi, Bridgers, Bortsova, Ghemawat, Ahmed, Liu, Powell, Bolina, Iinuma, Zablotskaia, Besley, Chung, Dozat, Comanescu, Si, Greer, Su, Polacek, Kaufman, Tokumine, Hu, Buchatskaya, Miao, Elhawaty, Siddhant, Tomasev, Xing, Greer, Miller, Ashraf, Roy, Zhang, Ma, Filos, Besta, Blevins, Klimenko, Yeh, Changpinyo, Mu, Chang,
  Pajarskas, Muir, Cohen, Lan, Haridasan, Marathe, Hansen, Douglas, Samuel, Wang, Austin, Lan, Jiang, Chiu, Lorenzo, Sjösund, Cevey, Gleicher, Avrahami, Boral, Srinivasan, Selo, May, Aisopos, Hussenot, Soares, Baumli, Chang, Recasens, Caine, Pritzel, Pavetic, Pardo, Gergely, Frye, Ramasesh, Horgan, Badola, Kassner, Roy, Dyer, Campos, Tomala, Tang, Badawy, White, Mustafa, Lang, Jindal, Vikram, Gong, Caelles, Hemsley, Thornton, Feng, Stokowiec, Zheng, Thacker, Çağlar Ünlü, Zhang, Saleh, Svensson, Bileschi, Patil, Anand, Ring, Tsihlas, Vezer, Selvi, Shevlane, Rodriguez, Kwiatkowski, Daruki, Rong, Dafoe, FitzGerald, Gu-Lemberg, Khan, Hendricks, Pellat, Feinberg, Cobon-Kerr, Sainath, Rauh, Hashemi, Ives, Hasson, Noland, Cao, Byrd, Hou, Wang, Sottiaux, Paganini, Lespiau, Moufarek, Hassan, Shivakumar, van Amersfoort, Mandhane, Joshi, Goyal, Tung, Brock, Sheahan, Misra, Li, Rakićević, Dehghani, Liu, Mittal, Oh, Noury, Sezener, Huot, Lamm, Cao, Chen, Mudgal, Stella, Brooks, Vasudevan, Liu, Chain, Melinkeri,
  Cohen, Wang, Seymore, Zubkov, Goel, Yue, Krishnakumaran, Albert, Hurley, Sano, Mohananey, Joughin, Filonov, Kępa, Eldawy, Lim, Rishi, Badiezadegan, Bos, Chang, Jain, Padmanabhan, Puttagunta, Krishna, Baker, Kalb, Bedapudi, Kurzrok, Lei, Yu, Litvin, Zhou, Wu, Sobell, Siciliano, Papir, Neale, Bragagnolo, Toor, Chen, Anklin, Wang, Feng, Gholami, Ling, Liu, Walter, Moghaddam, Kishore, Adamek, Mercado, Mallinson, Wandekar, Cagle, Ofek, Garrido, Lombriser, Mukha, Sun, Mohammad, Matak, Qian, Peswani, Janus, Yuan, Schelin, David, Garg, He, Duzhyi, Älgmyr, Lottaz, Li, Yadav, Xu, Chinien, Shivanna, Chuklin, Li, Spadine, Wolfe, Mohamed, Das, Dai, He, von Dincklage, Upadhyay, Maurya, Chi, Krause, Salama, Rabinovitch, M, Selvan, Dektiarev, Ghiasi, Guven, Gupta, Liu, Sharma, Shtacher, Paul, Akerlund, Aubet, Huang, Zhu, Zhu, Teixeira, Fritze, Bertolini, Marinescu, Bölle, Paulus, Gupta, Latkar, Chang, Sanders, Wilson, Wu, Tan, Thiet, Doshi, Lall, Mishra, Chen, Luong, Benjamin, Lee, Andrejczuk, Rabiej, Ranjan, Styrc,
  Yin, Simon, Harriott, Bansal, Robsky, Bacon, Greene, Mirylenka, Zhou, Sarvana, Goyal, Andermatt, Siegler, Horn, Israel, Pongetti, Chen, Selvatici, Silva, Wang, Tolins, Guu, Yogev, Cai, Agostini, Shah, Nguyen, Donnaile, Pereira, Friso, Stambler, Kurzrok, Kuang, Romanikhin, Geller, Yan, Jang, Lee, Fica, Malmi, Tan, Banica, Balle, Pham, Huang, Avram, Shi, Singh, Hidey, Ahuja, Saxena, Dooley, Potharaju, O'Neill, Gokulchandran, Foley, Zhao, Dusenberry, Liu, Mehta, Kotikalapudi, Safranek-Shrader, Goodman, Kessinger, Globen, Kolhar, Gorgolewski, Ibrahim, Song, Eichenbaum, Brovelli, Potluri, Lahoti, Baetu, Ghorbani, Chen, Crawford, Pal, Sridhar, Gurita, Mujika, Petrovski, Cedoz, Li, Chen, Santo, Goyal, Punjabi, Kappaganthu, Kwak, LV, Velury, Choudhury, Hall, Shah, Figueira, Thomas, Lu, Zhou, Kumar, Jurdi, Chikkerur, Ma, Yu, Kwak, Ähdel, Rajayogam, Choma, Liu, Barua, Ji, Park, Hellendoorn, Bailey, Bilal, Zhou, Khatir, Sutton, Rzadkowski, Macintosh, Vij, Shagin, Medina, Liang, Zhou, Shah, Bi, Dankovics, Banga,
  Lehmann, Bredesen, Lin, Hoffmann, Lai, Chung, Yang, Balani, Bražinskas, Sozanschi, Hayes, Alcalde, Makarov, Chen, Stella, Snijders, Mandl, Kärrman, Nowak, Wu, Dyck, Vaidyanathan, R, Mallet, Rudominer, Johnston, Mittal, Udathu, Christensen, Verma, Irving, Santucci, Elsayed, Davoodi, Georgiev, Tenney, Hua, Cideron, Leurent, Alnahlawi, Georgescu, Wei, Zheng, Scandinaro, Jiang, Snoek, Sundararajan, Wang, Ontiveros, Karo, Cole, Rajashekhar, Tumeh, Ben-David, Jain, Uesato, Datta, Bunyan, Wu, Zhang, Stanczyk, Zhang, Steiner, Naskar, Azzam, Johnson, Paszke, Chiu, Elias, Mohiuddin, Muhammad, Miao, Lee, Vieillard, Park, Zhang, Stanway, Garmon, Karmarkar, Dong, Lee, Kumar, Zhou, Evens, Isaac, Irving, Loper, Fink, Arkatkar, Chen, Shafran, Petrychenko, Chen, Jia, Levskaya, Zhu, Grabowski, Mao, Magni, Yao, Snaider, Casagrande, Palmer, Suganthan, Castaño, Giannoumis, Kim, Rybiński, Sreevatsa, Prendki, Soergel, Goedeckemeyer, Gierke, Jafari, Gaba, Wiesner, Wright, Wei, Vashisht, Kulizhskaya, Hoover, Le, Li, Iwuanyanwu,
  Liu, Ramirez, Khorlin, Cui, LIN, Wu, Aguilar, Pallo, Chakladar, Perng, Abellan, Zhang, Dasgupta, Kushman, Penchev, Repina, Wu, van~der Weide, Ponnapalli, Kaplan, Simsa, Li, Dousse, Yang, Piper, Ie, Pasumarthi, Lintz, Vijayakumar, Andor, Valenzuela, Lui, Paduraru, Peng, Lee, Zhang, Greene, Nguyen, Kurylowicz, Hardin, Dixon, Janzer, Choo, Feng, Zhang, Singhal, Du, McKinnon, Antropova, Bolukbasi, Keller, Reid, Finchelstein, Raad, Crocker, Hawkins, Dadashi, Gaffney, Franko, Bulanova, Leblond, Chung, Askham, Cobo, Xu, Fischer, Xu, Sorokin, Alberti, Lin, Evans, Dimitriev, Forbes, Banarse, Tung, Omernick, Bishop, Sterneck, Jain, Xia, Amid, Piccinno, Wang, Banzal, Mankowitz, Polozov, Krakovna, Brown, Bateni, Duan, Firoiu, Thotakuri, Natan, Geist, tan Girgin, Li, Ye, Roval, Tojo, Kwong, Lee-Thorp, Yew, Sinopalnikov, Ramos, Mellor, Sharma, Wu, Miller, Sonnerat, Vnukov, Greig, Beattie, Caveness, Bai, Eisenschlos, Korchemniy, Tsai, Jasarevic, Kong, Dao, Zheng, Liu, Yang, Zhu, Teh, Sanmiya, Gladchenko, Trdin, Toyama,
  Rosen, Tavakkol, Xue, Elkind, Woodman, Carpenter, Papamakarios, Kemp, Kafle, Grunina, Sinha, Talbert, Wu, Owusu-Afriyie, Du, Thornton, Pont-Tuset, Narayana, Li, Fatehi, Wieting, Ajmeri, Uria, Ko, Knight, Héliou, Niu, Gu, Pang, Li, Levine, Stolovich, Santamaria-Fernandez, Goenka, Yustalim, Strudel, Elqursh, Deck, Lee, Li, Levin, Hoffmann, Holtmann-Rice, Bachem, Arora, Koh, Yeganeh, Põder, Tariq, Sun, Ionita, Seyedhosseini, Tafti, Liu, Gulati, Liu, Ye, Chrzaszcz, Wang, Sethi, Li, Brown, Singh, Fan, Parisi, Stanton, Koverkathu, Choquette-Choo, Li, Lu, Ittycheriah, Shroff, Varadarajan, Bahargam, Willoughby, Gaddy, Desjardins, Cornero, Robenek, Mittal, Albrecht, Shenoy, Moiseev, Jacobsson, Ghaffarkhah, Rivière, Walton, Crepy, Parrish, Zhou, Farabet, Radebaugh, Srinivasan, van~der Salm, Fidjeland, Scellato, Latorre-Chimoto, Klimczak-Plucińska, Bridson, de~Cesare, Hudson, Mendolicchio, Walker, Morris, Mauger, Guseynov, Reid, Odoom, Loher, Cotruta, Yenugula, Grewe, Petrushkina, Duerig, Sanchez, Yadlowsky, Shen,
  Globerson, Webb, Dua, Li, Bhupatiraju, Hurt, Qureshi, Agarwal, Shani, Eyal, Khare, Belle, Wang, Tekur, Kale, Wei, Sang, Saeta, Liechty, Sun, Zhao, Lee, Nayak, Fritz, Vuyyuru, Aslanides, Vyas, Wicke, Ma, Eltyshev, Martin, Cate, Manyika, Amiri, Kim, Xiong, Kang, Luisier, Tripuraneni, Madras, Guo, Waters, Wang, Ainslie, Baldridge, Zhang, Pruthi, Bauer, Yang, Mansour, Gelman, Xu, Polovets, Liu, Cai, Chen, Sheng, Xue, Ozair, Angermueller, Li, Sinha, Wang, Wiesinger, Koukoumidis, Tian, Iyer, Gurumurthy, Goldenson, Shah, Blake, Yu, Urbanowicz, Palomaki, Fernando, Durden, Mehta, Momchev, Rahimtoroghi, Georgaki, Raul, Ruder, Redshaw, Lee, Zhou, Jalan, Li, Hechtman, Schuh, Nasr, Milan, Mikulik, Franco, Green, Nguyen, Kelley, Mahendru, Hu, Howland, Vargas, Hui, Bansal, Rao, Ghiya, Wang, Ye, Sarr, Preston, Elish, Li, Kaku, Gupta, Pasupat, Juan, Someswar, M., Chen, Amini, Fabrikant, Chu, Dong, Muthal, Buthpitiya, Jauhari, Hua, Khandelwal, Hitron, Ren, Rinaldi, Drath, Dabush, Jiang, Godhia, Sachs, Chen, Fan, Taitelbaum,
  Noga, Dai, Wang, Liang, Hamer, Ferng, Elkind, Atias, Lee, Listík, Carlen, van~de Kerkhof, Pikus, Zaher, Müller, Zykova, Stefanec, Gatsko, Hirnschall, Sethi, Xu, Ahuja, Tsai, Stefanoiu, Feng, Dhandhania, Katyal, Gupta, Parulekar, Pitta, Zhao, Bhatia, Bhavnani, Alhadlaq, Li, Danenberg, Tu, Pine, Filippova, Ghosh, Limonchik, Urala, Lanka, Clive, Sun, Li, Wu, Hongtongsak, Li, Thakkar, Omarov, Majmundar, Alverson, Kucharski, Patel, Jain, Zabelin, Pelagatti, Kohli, Kumar, Kim, Sankar, Shah, Ramachandruni, Zeng, Bariach, Weidinger, Vu, Andreev, He, Hui, Kashem, Subramanya, Hsiao, Hassabis, Kavukcuoglu, Sadovsky, Le, Strohman, Wu, Petrov, Dean, and Vinyals}]{geminiteam2025geminifamilyhighlycapable}
Gemini Team, Rohan Anil, Sebastian Borgeaud, Jean-Baptiste Alayrac, Jiahui Yu, Radu Soricut, Johan Schalkwyk, Andrew~M. Dai, Anja Hauth, Katie Millican, David Silver, Melvin Johnson, Ioannis Antonoglou, Julian Schrittwieser, Amelia Glaese, Jilin Chen, Emily Pitler, Timothy Lillicrap, Angeliki Lazaridou, and 1332 others. 2025{\natexlab{a}}.
\newblock \href {https://arxiv.org/abs/2312.11805} {Gemini: A family of highly capable multimodal models}.
\newblock \emph{Preprint}, arXiv:2312.11805.

\bibitem[{Team(2025)}]{kimiteam2025kimik15}
Kimi Team. 2025.
\newblock \href {https://arxiv.org/abs/2501.12599} {Kimi k1.5: Scaling reinforcement learning with llms}.
\newblock \emph{CoRR}, abs/2501.12599.

\bibitem[{Team et~al.(2025{\natexlab{b}})Team, Du, Yin, Xing, Qu, Wang, Chen, Zhang, Du, Wei, Wang, Zhang, Du, Wang, Yuan, Lu, Li, Sung, Wei, Lai, Zhu, Ding, Hu, Yang, Zhang, Wu, Yao, Lu, Wang, Gao, Zheng, Li, Su, Wang, Deng, Qiu, Xie, Wang, Liu, Yan, Ouyang, Chen, Sui, Yu, Dong, Dong, Xu, Cheng, Gu, Zhou, Liu, Cao, Yu, Song, Bai, Song, He, Huang, Xu, Yuan, Yao, Wu, Zu, Zhou, Wang, Charles, Zhong, Li, Hu, Chen, Wang, Liu, Miao, Qin, Chen, Bao, Wang, Kang, Liu, Du, Wu, Wang, Yan, Zhou, Li, Jiang, Zhang, Yang, Huang, Huang, Zhao, Chen, and Lin}]{kimiteam2025kimivl}
Kimi Team, Angang Du, Bohong Yin, Bowei Xing, Bowen Qu, Bowen Wang, Cheng Chen, Chenlin Zhang, Chenzhuang Du, Chu Wei, Congcong Wang, Dehao Zhang, Dikang Du, Dongliang Wang, Enming Yuan, Enzhe Lu, Fang Li, Flood Sung, Guangda Wei, and 74 others. 2025{\natexlab{b}}.
\newblock \href {https://arxiv.org/abs/2504.07491} {Kimi-vl technical report}.

\bibitem[{Trinh et~al.(2024)Trinh, Wu, Le, He, and Luong}]{alphaGeometry24}
Trieu~H Trinh, Yuhuai Wu, Quoc~V Le, He~He, and Thang Luong. 2024.
\newblock Solving olympiad geometry without human demonstrations.
\newblock \emph{Nature}, 625(7995):476--482.

\bibitem[{Wu et~al.(2023)Wu, Bansal, Zhang, Wu, Zhang, Zhu, Li, Jiang, Zhang, and Wang}]{DBLP:journals/corr/autogen}
Qingyun Wu, Gagan Bansal, Jieyu Zhang, Yiran Wu, Shaokun Zhang, Erkang Zhu, Beibin Li, Li~Jiang, Xiaoyun Zhang, and Chi Wang. 2023.
\newblock \href {https://doi.org/10.48550/ARXIV.2308.08155} {Autogen: Enabling next-gen {LLM} applications via multi-agent conversation framework}.
\newblock \emph{CoRR}, abs/2308.08155.

\bibitem[{Xu et~al.(2025)Xu, Chan, Li, Aljunied, Yuan, Wang, Xiao, Chen, Liu, Li et~al.}]{xu2025lingshu}
Weiwen Xu, Hou~Pong Chan, Long Li, Mahani Aljunied, Ruifeng Yuan, Jianyu Wang, Chenghao Xiao, Guizhen Chen, Chaoqun Liu, Zhaodonghui Li, and 1 others. 2025.
\newblock Lingshu: A generalist foundation model for unified multimodal medical understanding and reasoning.
\newblock \emph{arXiv preprint arXiv:2506.07044}.

\bibitem[{Yang et~al.(2025)Yang, He, Pan, Jiang, Deng, Yang, Lu, Yin, Rao, Zhu, Zhang, and Chen}]{yang2025r1onevision}
Yi~Yang, Xiaoxuan He, Hongkun Pan, Xiyan Jiang, Yan Deng, Xingtao Yang, Haoyu Lu, Dacheng Yin, Fengyun Rao, Minfeng Zhu, Bo~Zhang, and Wei Chen. 2025.
\newblock R1-onevision: Advancing generalized multimodal reasoning through cross-modal formalization.
\newblock \emph{arXiv preprint arXiv:2503.10615}.

\bibitem[{Yuan et~al.(2025)Yuan, Xiao, Leng, Wang, Li, Xu, Chan, Zhao, Xu, Wei et~al.}]{yuan2025vl}
Ruifeng Yuan, Chenghao Xiao, Sicong Leng, Jianyu Wang, Long Li, Weiwen Xu, Hou~Pong Chan, Deli Zhao, Tingyang Xu, Zhongyu Wei, and 1 others. 2025.
\newblock Vl-cogito: Progressive curriculum reinforcement learning for advanced multimodal reasoning.
\newblock \emph{arXiv preprint arXiv:2507.22607}.

\bibitem[{Zhang et~al.(2024{\natexlab{a}})Zhang, Liu, Yu, Hu, and Neiswanger}]{zhang2024euclidsuperchargingmultimodalllms}
Jiarui Zhang, Ollie Liu, Tianyu Yu, Jinyi Hu, and Willie Neiswanger. 2024{\natexlab{a}}.
\newblock \href {https://arxiv.org/abs/2412.08737} {Euclid: Supercharging multimodal llms with synthetic high-fidelity visual descriptions}.
\newblock \emph{Preprint}, arXiv:2412.08737.

\bibitem[{Zhang et~al.(2024{\natexlab{b}})Zhang, Jiang, Zhang, Lin, Guo, Qiu, Zhou, Lu, Chang, Qiao et~al.}]{zhang2024mathverse}
Renrui Zhang, Dongzhi Jiang, Yichi Zhang, Haokun Lin, Ziyu Guo, Pengshuo Qiu, Aojun Zhou, Pan Lu, Kai-Wei Chang, Yu~Qiao, and 1 others. 2024{\natexlab{b}}.
\newblock \href {https://doi.org/10.1007/978-3-031-73242-3_10} {Mathverse: Does your multi-modal llm truly see the diagrams in visual math problems?}
\newblock In \emph{European Conference on Computer Vision}, pages 169--186, Berlin, Heidelberg. Springer-Verlag.

\bibitem[{Zhang et~al.(2025)Zhang, Wei, Jiang, Guo, Zhang, Tong, Liu, Zhou, Zhang, Gao, and Li}]{zhang2025mavis}
Renrui Zhang, Xinyu Wei, Dongzhi Jiang, Ziyu Guo, Yichi Zhang, Chengzhuo Tong, Jiaming Liu, Aojun Zhou, Shanghang Zhang, Peng Gao, and Hongsheng Li. 2025.
\newblock \href {https://openreview.net/forum?id=MnJzJ2gvuf} {{MAVIS}: Mathematical visual instruction tuning with an automatic data engine}.
\newblock In \emph{The Thirteenth International Conference on Learning Representations}.

\end{thebibliography}
\clearpage
\appendix

\section{Details of GeoPQA}~\label{appendix:GeoPQA}

\subsection{Prompt to generate perception QAs}\label{appendix:prompt}
The following prompt is used with Gemini-2.0-Flash-Thinking to generate the initial set of perception question-answer pairs.

\begin{mdframed}
[
  backgroundcolor=Gainsboro,
  linecolor=black,
  linewidth=1pt,
  roundcorner=3pt,
]
\fontsize{9pt}{10pt}\selectfont
\ttfamily
Create perception questions based on the provided image description. The questions should be formulated such that:\\
1. They involve recognizing basic visual elements and spatial relationships directly from the image.\\
2. They are answerable from the image description.\\
3. Answers must be "yes/no", a number, or a simple string like "AB" (no spaces).\\
4. No reasoning should be provided with the answer.\\
5. Avoid rephrasing the same question.\\
6. Output the results as a JSON array of objects. Each object should have keys "question" and "answer". If no meaningful question can be generated, return an empty array. If the image is too simple, for example, only contains a single point or a line segment, return an empty array.\\
7. No more than seven questions should be generated.\\\\
Image Description: Triangle ABC is a right angle isosceles triangle, with $\angle$BAC as the right angle. The circle that passes through points A, C, and B has center D.\\
Questions: [\{"question": "Is triangle ABC a right triangle?", "answer": "Yes"\}, \{"question": "Which vertex has the right angle in triangle ABC?", "answer": "A"\}, \{"question": "Does the circle pass through point D?", "answer": "No"\}, \{"question": "What is the measure of angle BAC?", "answer": "90"\}, \{"question": "Are sides AB and AC equal in length?", "answer": "Yes"\}]\\\\
Image Description: \textcolor{purple}{<description>}\\
Questions:
\end{mdframed}

\subsection{Synthetic geometric diagram generation}\label{appendix:synthetic}

We create the following to generate the synthetic geometric diagrams: 
\begin{itemize}
    \item \textbf{Basic shapes}: Line segments, circles, triangles, quadrilaterals, and pentagons.
    \item \textbf{Composite shapes}: Combinations of 2-4 random basic shapes with predefined spatial relationships (e.g., a circle tangent to a triangle).
    \item \textbf{Annotations}: Diagrams are explicitly annotated with special geometric symbols, such as right-angle symbols and markings for equal sides/angles, which are commonly understood by humans but potentially ambiguous for MLLMs.
\end{itemize}

\subsection{Dataset statistics}\label{appendix:data-stats}

The dataset is split into 659 test samples and 5420 training samples. The training set contains a balanced mix of real-world and synthetic images. The distribution is shown in Table \ref{tab:real-syn}.

\begin{table}[ht] 
  \centering
  \resizebox{0.7\linewidth}{!}{
  \begin{tabular}{@{}lcc@{}}
    \toprule
     \textbf{Image Type} & \textbf{\# Images} & \textbf{\# Questions} \\
    \midrule
    Real               & 2548 & 7038 \\
    Synthetic &	2872 &	9303 \\
    \bottomrule
  \end{tabular}
  }
\caption{Distribution of real vs. synthetic images.}
  \label{tab:real-syn} 
\end{table}

To provide an estimate of sample complexity, we analyse the number of perception questions associated with each image, which serves as a proxy for its visual complexity. The distribution is shown in Table \ref{tab:question-num}.

\begin{table}[ht] 
  \centering
  \resizebox{0.6\linewidth}{!}{
  \begin{tabular}{@{}cc@{}}
    \toprule
     \textbf{\# Questions per Sample} & \textbf{\% Share} \\
    \midrule
    1	& 9.56 \\
    2	& 23.28 \\
    3   & 35.15 \\
    4   & 22.38 \\
    5+	& 9.63 \\
    \bottomrule
  \end{tabular}
  }
\caption{Percentage share of the number of questions per sample.}
  \label{tab:question-num} 
\end{table}

The created perception questions cover a range of geometric concepts, including (1) \textbf{basic geometric elements} such as identifying shapes (\eg triangles, circles), comparing lengths and recognising angles (\eg right, acute, obtuse); (2) \textbf{geometric relationships} such as intersection, parallelism, perpendicularity, and tangency. Table \ref{tab:question-type}
shows the distribution of question types in GeoPQA.

\begin{table}[ht]
  \centering
  \resizebox{\linewidth}{!}{
  \begin{tabular}{@{}llr@{}}
    \toprule
    \textbf{Category} & \textbf{Sub-category} & \textbf{Count} \\
    \midrule
    \multirow{5}{*}{Geometric Elements} & Shapes & 4387 \\
    & Angles & 1737 \\
    & Lengths & 1405 \\
    & Area/Perimeter & 46 \\
    & Others & 243 \\
    \midrule
    \multirow{7}{*}{Geometric Relationships} & Relative Position & 5662 \\
    & Intersection & 1108 \\
    & Perpendicularity & 500 \\
    & Parallelism & 234 \\
    & Tangency & 432 \\
    & Congruence/Similarity & 410 \\
    & Transformation & 177 \\
    \bottomrule
  \end{tabular}
  }
  \caption{Distribution of question types.}
  \label{tab:question-type} 
\end{table}

\subsection{Quality control of generated perception QAs}
\label{appendix:quality}

We prompt GPT-4o~\citep{openai2024gpt4o} to filter out low-quality questions. The following prompt is used.

\begin{mdframed}
[
  backgroundcolor=Gainsboro,
  linecolor=black,
  linewidth=1pt,
  roundcorner=3pt,
]
\fontsize{9pt}{10pt}\selectfont
\ttfamily
Your task is to evaluate the correctness of a user's answer based on an image, its description, and a given question. The user's answer is considered incorrect if:\\
- The image does not explicitly contain the information needed to answer the question.\\
- The answer contradicts the information presented in the image description.\\\\
**Input**:\\
- Image Description: \textcolor{purple}{<description>}\\
- Question: \textcolor{purple}{<question>}\\
- User's Answer: \textcolor{purple}{<response>}\\\\
**Output Format**:\\
Provide your reasoning and judgment (0 = correct, 1 = incorrect) in the following format:\\
<think>\{\{Your concise reasoning, including consideration of the image description and question, and how it relates to the user's answer.\}\}</think>
<judge>\{\{0 or 1, 0 if the user's answer is correct, 1 if the user's answer is incorrect.\}\}</judge>
\end{mdframed}

To validate the quality of the dataset after GPT-4o filtering, we perform a human inspection on 100 random samples: 92\% are valid and high-quality. The 8\% invalid samples comprised 2\% from the synthetic subset and 6\% from the real-world image subset. While there is a slight error rate, our main results show that perception training on this dataset still yields a significant benefit over reasoning-only training. This demonstrates the practical effectiveness of our dataset. Furthermore, since most errors are from the real-world subset, we can further improve the dataset quality by increasing the proportion of high-quality synthetic data in future iterations.




\section{Training setup}\label{appendix:training}

Table \ref{tab:hyperparameters} shows the hyperparameter configuration in our training. We adopt the same settings across all experiments to ensure a fair and direct comparison.

\begin{table}[ht] 
  \centering
  \resizebox{0.7\linewidth}{!}{
  \begin{tabular}{@{}ll@{}}
    \toprule
     \textbf{Hyperparameter} & \textbf{Configuration} \\
    \midrule
    Max Prompt Length    & 2048 \\
    Max Response Length  & 2048 \\
    Max Image Pixels     & 1,048,576 \\
    Min Image Pixels     & 65,536 \\
    Global Batch Size    & 128 \\
    Rollout Batch Size   & 512 \\
    Learning Rate        & 1e-6 \\
    Optimizer            & AdamW \\
    N Rollouts           & 5 \\
    Training Episodes    & 10 \\
    \bottomrule
  \end{tabular}
  }
  \caption{Hyperparameters used in training.}
  \label{tab:hyperparameters} 
\end{table}

\section{Evaluation}\label{appendix:evaluation}
All evaluation is conducted using the VLMEvalKit toolkit\footnote{\url{https://github.com/open-compass/VLMEvalKit}}, ensuring standardized and reproducible evaluation metrics.

\section{Performance on other tasks}
\label{sec:othertasks}
Table \ref{tab:mathvista-other-tasks} shows the performance of our method vs. the reasoning-only method on other MathVista tasks.

\begin{table}[ht] 
  \centering

  \resizebox{\linewidth}{!}{
  \begin{tabular}{@{}lcc@{}}
    \toprule
    MathVista Task Category & Reasoning-only & Perception + Reasoning \\
    \midrule
    Figure Question Answering               & 68.0   & 69.5  \\
    Textbook Question Answering             & 62.0   & 64.6   \\
    VisualQA                & 58.1   & 57.0  \\
    Scientific Reasoning    & 59.8   & 62.3  \\
    Numeric Commonsense     & 43.1   & 40.3   \\
    Arithmetic Reasoning    & 56.7   & 58.1 \\
    Algebraic Reasoning     & 62.6   & 69.4 \\
    Math word problem       & 62.9   & 61.8  \\
    Logical Reasoning       & 29.7   & 37.8 \\
    \bottomrule
  \end{tabular}
  }
\caption{Performance (\%) on MathVista other tasks. ``Perception + Reasoning'' refers to our two-stage approach.}
  \label{tab:mathvista-other-tasks} 
\end{table}

\end{document}